\documentclass{article}
\usepackage{ijcai20}

\usepackage{times}
\usepackage{soul}
\usepackage{url}
\usepackage[hidelinks]{hyperref}
\usepackage[utf8]{inputenc}
\usepackage[small]{caption}
\usepackage{graphicx}
\usepackage{amsmath}
\usepackage{amsthm}
\usepackage{booktabs}
\usepackage{algorithm}
\usepackage{algorithmic}
\usepackage{amsmath, amssymb, amsthm, amsfonts}
\usepackage[multiple]{footmisc}
\usepackage{multirow}
\usepackage{subcaption}
\usepackage{xcolor}

\newtheorem{definition}{Definition}

\title{Generating Higher-Fidelity Synthetic Datasets with Privacy Guarantees}

\author{
Aleksei Triastcyn
\and
Boi Faltings
\affiliations
Artificial Intelligence Lab\\
EPFL, Switzerland\\
\emails
\{aleksei.triastcyn, boi.faltings\}@epfl.ch,
}

\begin{document}

\maketitle

\begin{abstract}
This paper considers the problem of enhancing user privacy in common machine learning development tasks, such as data annotation and inspection, by substituting the real data with samples form a generative adversarial network. We propose employing Bayesian differential privacy as the means to achieve a rigorous theoretical guarantee while providing a better privacy-utility trade-off. We demonstrate experimentally that our approach produces higher-fidelity samples, compared to prior work, allowing to (1) detect more subtle data errors and biases, and (2) reduce the need for real data labelling by achieving high accuracy when training directly on artificial samples.
\end{abstract}

\section{Introduction}
\label{sec:introduction}
With machine learning (ML) becoming ubiquitous in many aspects of our society, questions of its privacy and security take centre stage. A growing field of research in privacy attacks on ML~\cite{fredrikson2015model,shokri2017membership,hitaj2017deep,truex2018towards} tells us that it is possible to infer information about training data even in a black-box setting, without access to model parameters. A wider population, however, is concerned with privacy practices used in the ML development cycle, such as company employees or contractors manually inspecting and annotating user data\footnote{\url{https://www.theguardian.com/technology/2020/jan/10/skype-audio-graded-by-workers-in-china-with-no-security-measures}}\footnote{\url{https://www.bloomberg.com/news/articles/2019-04-10/is-anyone-listening-to-you-on-alexa-a-global-team-reviews-audio}}.

The problem of privacy attacks is often tackled with adding a differentially private mechanism to the model training procedure~\cite{abadi2016deep}. Differential privacy (DP)~\cite{dwork2006} provides a rigorous theoretical guarantee, which states (informally) that the algorithm output would not significantly change when a single user adds or removes their data, except with small (failure) probability. Another approach gaining popularity is \emph{federated learning} (FL)~\cite{mcmahan2016communication,bonawitz2017practical}, where a central entity trains a model by computing updates locally on-device and then securely aggregating these updates on a server. This way user data never leave their devices.

In spite of significant progress, neither of these approaches solves the problem of manual data labelling. Moreover, it creates an additional hurdle for developers, as they cannot inspect data, especially in decentralised settings, making it difficult to understand the model behaviour and find bugs in data and implementations. \citeauthor{augenstein2019generative}~\shortcite{augenstein2019generative} provide a more complete characterisation of these questions.

This paper follows~\citeauthor{augenstein2019generative}~\shortcite{augenstein2019generative} in adopting generative adversarial networks (GAN)~\cite{goodfellow2014generative} trained in a privacy-preserving manner for addressing these issues. More specifically, we use the notion of Bayesian differential privacy (BDP)~\cite{triastcyn2019bayesian}, which takes into account the data distribution and provides a more meaningful guarantee for in-distribution samples than classical DP. Intuitively, when DP has uniform failure probability for all data points, BDP allows it to be non-uniform, thereby discounting points that are naturally difficult to hide and providing a strong guarantee for the rest of the dataset. Since both can use the same obfuscation mechanism, while computing two privacy bounds in parallel, a DP guarantee would still hold for out-of-distribution samples. More details on the overall approach and privacy are provided in Section~\ref{sec:approach}.

The advantage of using this privacy definition is that it enables generating data of higher fidelity, compared to previous work on GANs with DP, allowing for finer-grained inspection of data. While some problems with data or data pipelines can be discovered using very coarse samples (e.g. pixel intensity inversion in~\cite{augenstein2019generative}), more subtle bugs, like partial data corruption, would require samples of much better quality, rendering the DP guarantee too loose to be meaningful. Moreover, if fidelity is high enough, synthetic data can be used for annotation and training itself, removing the related privacy concerns and extending applicability of FL. We evaluate our solution in these two aspects in Section~\ref{sec:evaluation}.

The main contributions of this paper are as follows:
\begin{itemize}
\item we use Bayesian DP to enable higher quality GAN samples, while still providing a strong privacy guarantee;
\item we demonstrate that this technique can be used to discover finer data errors than has been previously reported;
\item we also show that for some tasks synthetic data are of high enough quality to be used for labelling and training.
\end{itemize}

\section{Related Work}
\label{sec:related_work}
A rapidly expanding area of privacy-preserving machine learning research has been recently focused on the attacks that compromise privacy of training data, such as model inversion~\cite{fredrikson2015model} and membership inference~\cite{shokri2017membership}. The former is based on observing the output probabilities of the target model for a given class and performing gradient descent on an input reconstruction. The latter assumes an attacker with access to similar data, which is used to train "shadow" models, mimicking the target, and the attack model, which predicts if a certain example has already been seen during training based on its output probabilities. Both attacks can be performed in a black-box setting, without access to the model internal parameters.

Differential privacy (DP)~\cite{dwork2006} is widely accepted as the gold standard for preventing such attacks. One of the early takes on the problem is to use disjoint datasets and distributed training with DP. For example,  \cite{shokri2015privacy} propose to train a model in a distributed manner by communicating sanitised updates from participants to a central authority. Such a method, however, yields high privacy losses~\cite{abadi2016deep,papernot2016semi}. An alternative technique suggested by \cite{papernot2016semi} also uses disjoint training sets and builds an ensemble of independently trained teacher models to transfer knowledge to a student model by labelling public data. This result has been extended in \cite{papernot2018scalable} to achieve state-of-the-art image classification results in a private setting (with single-digit DP bounds).
A different approach is taken by \cite{abadi2016deep}. They propose using differentially private stochastic gradient descent (DP-SGD) to train deep learning models in a private manner. This approach achieves high accuracy maintaining relatively low DP bounds and being simpler to implement, but may also require pre-training on public data.

Due to the fact that the DP threat model is extremely broad, achieving a reasonable guarantee may be difficult or even impossible. For this reason, a number of alternative definitions has been proposed over the recent years, aimed at relaxing the guarantee or providing tighter composition bounds under certain assumptions. Examples are computational DP~\cite{mironov2009computational}, mutual-information privacy~\cite{mir2012information,wang2016relation}, different versions of concentrated DP (CDP~\cite{dwork2016concentrated}, zCDP~\cite{bun2016concentrated}, tCDP~\cite{bun2018composable}), and R\'enyiDP (RDP)~\cite{mironov2017renyi}. Some other relaxations~\cite{abowd2013differential,schneider2015new,charest2017meaning} tip the balance even further in favour of applicability at the cost of weaker guarantees, for example considering the average-case instead of the worst-case~\cite{triastcyn2019federated}.

In this work, we rely on another relaxation, called Bayesian differential privacy~\cite{triastcyn2019bayesian}. This notion utilises the fact that data come from a particular distribution, and not all data samples are equally likely (e.g. unlikely to find a sound record among ECG samples). At the same time, it maintains a similar probabilistic interpretation of its parameters $\varepsilon$ and $\delta$. It is worth noting, that unlike some of the relaxations mentioned above, Bayesian DP can provide a tail bound on privacy loss, similarly to the moments accountant (MA)~\cite{abadi2016deep}, and is not limited to a particular dataset, but rather a particular type of data (e.g. emails, MRI images, etc.), which is a much more permitting assumption.

Up until recently, another aspect of privacy in machine learning has been largely overlooked: the human involvement in the development cycle and manual data processing. These issues can be mitigated, at least partially, by federated learning (FL)~\cite{mcmahan2016communication}, which brings a great promise for user privacy. Yet, FL paradigm creates additional problems of its own. \citeauthor{augenstein2019generative}~\shortcite{augenstein2019generative} provide a good starting point, systematising these problems and proposing a solution by the use of synthetic data. Although privacy-preserving data synthesis using GANs has been introduced in earlier works~\cite{beaulieu2017privacy,xie2018differentially,zhang2018differentially,triastcyn2019federated,jordon2018pate,long2019scalable}, these papers mainly focused on achieving high utility of synthetic data without addressing a broader scope of privacy leakage via manual data handling.

A common problem of privacy-preserving GANs, however, is that the generated samples have very low fidelity, unless the privacy guarantee is unreasonably weak. Our approach makes progress in exactly this perspective: we can achieve much higher quality outputs with little compromise in privacy guarantees (and only for outliers that are difficult to hide). As a result, our synthetic data yield better performance of downstream analytics, and simultaneously, provide more powerful data inspection capabilities.

\section{Preliminaries}
\label{sec:preliminaries}
In this section, we provide some background useful for understanding the paper.

We use $D, D'$ to represent neighbouring (adjacent) datasets. If not specified, it is assumed that these datasets differ in a single example. Individual examples in a dataset are denoted by $x$ or $x_i$, while the example by which two datasets differ---by $x'$. We assume $D' = D \cup \{x'\}$, whenever possible to do so without loss of generality. The private learning outcomes (i.e. noised gradients) are denoted by $w$.

\begin{definition}
A randomised function (mechanism) $\mathcal{A}: \mathcal{D} \rightarrow \mathcal{R}$ with domain $\mathcal{D}$ and range $\mathcal{R}$ satisfies $(\varepsilon, \delta)$-differential privacy if for any two adjacent inputs $D, D' \in \mathcal{D}$ and for any set of outcomes $\mathcal{S} \subset \mathcal{R}$ the following holds:
\begin{align}
	\Pr\left[\mathcal{A}(D) \in \mathcal{S}\right] \leq e^\varepsilon \Pr\left[\mathcal{A}(D') \in \mathcal{S}\right] + \delta.
\end{align}
\end{definition}

\begin{definition}
Privacy loss of a randomised mechanism $\mathcal{A}: \mathcal{D} \rightarrow \mathcal{R}$ for inputs $D, D' \in \mathcal{D}$ and outcome $w \in \mathcal{R}$ takes the following form:
\begin{align}
	L(w, D, D') = \log\frac{\Pr\left[\mathcal{A}(D) = w \right]}{\Pr\left[\mathcal{A}(D') = w \right]}.
\end{align}
\end{definition}

\begin{definition}
The Gaussian noise mechanism achieving $(\varepsilon, \delta)$-DP, for a function $f: \mathcal{D} \rightarrow \mathbb{R}^m$, is defined as
\begin{align}
	\mathcal{A}(D) = f(D) + \mathcal{N}(0, I\sigma^2),
\end{align}
where $\sigma > C \sqrt{2\log\frac{1.25}{\delta}} / \varepsilon$ and $C$ is the L2-sensitivity of $f$.
\end{definition}

For more details on differential privacy, the Gaussian mechanism, and how to use it in machine learning, we refer the reader to~\cite{dwork2014algorithmic,abadi2016deep}.

\begin{definition}
\label{def:bayes_dp}
A randomised function (algorithm) $\mathcal{A}: \mathcal{D} \rightarrow \mathcal{R}$ with domain $\mathcal{D}$ and range $\mathcal{R}$ satisfies $(\varepsilon_\mu, \delta_\mu)$-(weak) Bayesian differential privacy if for any two adjacent datasets $D, D' \in \mathcal{D}$, \emph{differing in a single data point $x' \sim \mu(x)$}, and for any set of outcomes $\mathcal{S} \subset \mathcal{R}$ the following holds:
\begin{align}
	\Pr\left[\mathcal{A}(D) \in \mathcal{S} \right] \leq e^{\varepsilon_\mu} \Pr\left[\mathcal{A}(D') \in \mathcal{S} \right] + \delta_\mu.
\end{align}
\end{definition}

While the definition of BDP is very close to that of DP, there are some important differences: the interpretation of $\delta$ is slightly different, data are assumed to come from a distribution $\mu(x)$ (although it is not required to be known), and samples are assumed to be exchangeable~\cite{aldous1985exchangeability}. Nonetheless, this notion remains applicable in a wide range of practical scenarios~\cite{triastcyn2019bayesian}.

In parts of the paper, we refer to \citeauthor{augenstein2019generative} classification of ML developer tasks, which can be condensed to:

T1 - Sanity checking data.

T2 - Debugging mistakes.

T3 - Debugging unknown labels / classes.

T4 - Debugging poor performance on certain classes / slices / users.

T5 - Human labelling of examples.

T6 - Detecting bias in the training data.

\section{Our Approach}
\label{sec:approach}
In this section, we describe our approach, intuition behind it, its privacy analysis, and discuss how to extend it to federated learning settings.

\subsection{Intuition}
\label{sec:intuition}
The primary distinction of Bayesian differential privacy is that it takes into account the data distribution, and by extension, assumes that all data points are drawn from the same distribution, although these distributions may be multimodal, highly complex, and generally unknown. This is a natural hypothesis in many machine learning applications, but especially so when working with generative models like GANs.

The task of generative modelling in itself is to learn an underlying data distribution, and thus, a common distribution is an implicit belief. This results in an organic match with BDP, because there are no assumptions to add to the problem.

Another part of our intuition is that the foremost source of privacy leakage are outliers. On the one hand, their respective privacy loss would be discounted in BDP accounting due to their low probability. On the other hand, we can reduce the number of samples generated by the GAN to decrease the chances of these outliers appearing in the synthetic dataset.

\subsection{Overview}
\label{sec:overview}
We are given a dataset $D$ of labelled ($\{ (x_i, y_i) ~|~ (x_i, y_i) \sim \mu(x, y),~ i = 1..n \}$) or unlabelled ($\{ x_i ~|~ x_i \sim \mu(x),~ i = 1..n \}$) examples. This dataset can be decentralised, in which case we would use FL (see Section~\ref{sec:federated}). Our task is to train a GAN, which consists of the generator $\mathcal{G}$ and the critic $\mathcal{C}$ (discriminator), to generate synthetic samples from $\mu$.

Our privacy mechanism follows the previous work on privacy-preserving GANs~\cite{beaulieu2017privacy,xie2018differentially}. More specifically, it applies the Gaussian mechanism (clip to norm $C$ and add Gaussian noise with variance $C^2\sigma^2$) to discriminator updates at each step of the training. Privacy of the generator is then guaranteed by the post-processing property of BDP. It is worth mentioning, however, that clipping and/or adding noise to generator gradients can be beneficial for training in some cases, to keep a better balance in the game between the critic and the generator, and it should not be overlooked by developers.

We choose not to implement more complicated schemes, such as PATE-GAN~\cite{jordon2018pate} or G-PATE~\cite{long2019scalable}, which use PATE framework~\cite{papernot2018scalable} to guarantee differential privacy for GANs. Our key rationale is that a more complicated structure of this solution could create unnecessary errors and additional privacy leakage (e.g. leaking privacy by backpropagating through the teachers' votes to the generator, thereby neglecting the added noise). Nevertheless, we show in our evaluation that due to the distribution-calibrated BDP accounting (and hence, less added noise) our GAN generates better quality samples compared to these more complex solutions.

\subsection{Privacy Analysis}
\label{sec:privacy}
In order to compute privacy guarantees of the synthetic dataset w.r.t. the real one, we need to bound privacy loss of the generative model. As noted before, we effectively enforce privacy on the critic and then rely on preservation of guarantees under post-processing. This arrangement ensures a simple adoption of privacy accounting for discriminative models.

Privacy accounting is done by using the Bayesian accountant~\cite{triastcyn2019bayesian}. To benefit from the data distribution information, it needs to sample a number of gradients at each iteration in addition to the one used in the update. These gradients are then used to estimate the upper confidence bound on the privacy cost $c_t(\lambda)$:
\begin{align}
c_t(\lambda) = \max\{c_t^{L}(\lambda), c_t^{R}(\lambda)\},
\end{align}
where
\begin{align}
	&c_t^{L}(\lambda) = \log \mathbb{E}_x \left[ \mathbb{E}_{k \sim B(\lambda+1, q)} \left[e^{\frac{k^2 - k}{2\sigma^2} \|g_t - g'_t\|^2} \right] \right], \\
	&c_t^{R}(\lambda) = \log \mathbb{E}_x \left[ \mathbb{E}_{k \sim B(\lambda, q)} \left[e^{\frac{k^2 + k}{2\sigma^2} \|g_t - g'_t\|^2} \right] \right].
\end{align}
Here, $B(\lambda, q)$ is the binomial distribution with $\lambda$ experiments (a hyper-parameter) and the probability of success $q$ (equal to the probability of sampling a single data point in a batch), $g_t$ and $g'_t$ are two gradient samples differing in one data point.

The privacy guarantee is calculated from the privacy cost, by fixing either $\varepsilon_\mu$ or $\delta_\mu$:
\begin{align}
	\log \delta_\mu \leq \sum_{t=1}^T c_t(\lambda) - \lambda \varepsilon_\mu.
\end{align}
For more details on the Bayesian accountant and related proofs, see~\cite{triastcyn2019bayesian}.

An important difference in privacy accounting for GANs is that not every update of the critic should be accounted for. Updates on fake data samples do not leak information about the real data beyond what is already accounted for in the previous iterations. Therefore, only real updates are sampled and used for the privacy cost estimation. In some GAN architectures, however, one should be careful to consider additional sources of privacy leakage, such as the gradient penalty in WGAN-GP~\cite{gulrajani2017improved}.

To better understand how the BDP bound relates to the traditional DP, consider the following conditional probability:
\begin{align}
\Delta(\varepsilon, x') = \Pr\left[ L(w, D, D') > \varepsilon ~|~ D, D' = D \cup \{x'\} \right].
\end{align}

The moments accountant outputs $\delta$ that upper-bounds $\Delta(\varepsilon, x')$ for all $x'$. It is not true in general for other accounting methods, but let us focus on MA, as it is by far the most popular. Consequently, the moments accountant bound is
\begin{align}
\max_{x} \Delta(\varepsilon, x) \leq \delta,
\end{align}
where $\varepsilon$ is a chosen constant. At the same time, BDP bounds the probability that is not conditioned on $x'$, but we can transform one to another through marginalisation and obtain:
\begin{align}
\mathbb{E}_{x} \left[ \Delta(\varepsilon, x) \right] \leq \delta_\mu.
\end{align}
On the surface, this guarantee seems considerably weaker, as it holds only in expectation. However, since $\Delta(\cdot)$ is a non-negative random variable in $x$, we can apply Markov's inequality and obtain a tail bound on it using $\delta_\mu$. \emph{We can therefore find a pair $(\varepsilon, \delta)_p$ that holds for any percentile $p$ of the data/user distribution, not just in expectation.} In all our experiments, we consider bounds well above 99th percentile, so it is very unlikely to encounter data for which the equivalent DP guarantee doesn't hold.

\begin{figure}
	\centering
	\begin{subfigure}{0.49\linewidth}
    		\includegraphics[width=\textwidth]{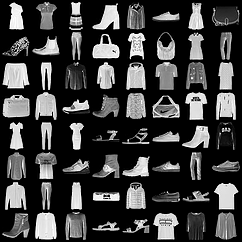}
    		\caption{Real.}
	\end{subfigure}
	\begin{subfigure}{0.49\linewidth}
    		\includegraphics[width=\textwidth]{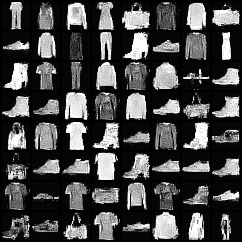}
    		\caption{Synthetic.}
	\end{subfigure}
	\caption{Real and synthetic samples on Fashion-MNIST.}
	\label{fig:fashion_real_fake}
\end{figure}

\subsection{Federated Learning Case}
\label{sec:federated}
In the above description, we did not make any assumptions on where the data are located. The most logical scenario to consider is federated learning, like in~\cite{augenstein2019generative}, such that the data remain on user devices at all times.

To accommodate FL scenarios, minimal modifications to the approach are required. Training of the generative model would be performed in the same way as any other federated model, and privacy accounting would be done at the user-level~\cite{augenstein2019generative}. Baysian DP results are also directly transferable to FL~\cite{triastcyn2019federatedbdp}, and privacy bounds are generally even tighter in this case.

\section{Evaluation}
\label{sec:evaluation}
In this section, we describe the experimental setup, implementation, and evaluate our method on MNIST~\cite{lecun1998gradient} and Fashion-MNIST~\cite{xiao2017online} datasets.

\subsection{Experimental Setting}
\label{sec:setting}

\begin{figure}
	\centering
	\begin{subfigure}{0.32\linewidth}
    		\includegraphics[width=\textwidth]{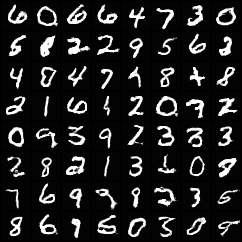}
    		\caption{Trained on correct images with BDP.}
    		\label{fig:debug_mnist_bdp_correct}
	\end{subfigure}
	\begin{subfigure}{0.32\linewidth}
    		\includegraphics[width=\textwidth]{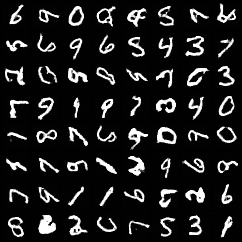}
    		\caption{Trained on altered images with BDP.}
    		\label{fig:debug_mnist_bdp}
	\end{subfigure}
	\begin{subfigure}{0.32\linewidth}
    		\includegraphics[width=\textwidth]{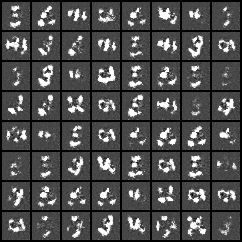}
    		\caption{Trained on altered images with DP.}
    		\label{fig:debug_mnist_dp}
	\end{subfigure}
	\caption{GAN output for detecting unwanted rotations on MNIST.}
	\label{fig:debug_mnist}
\end{figure}

\begin{table*}
	\caption{Accuracy of models: (1) non-private baseline (convolutional network); (2) private classifier (convolutional network trained with BDP); and student models: (3) for G-PATE with $(1, 10^{-5})$-DP guarantee; (4) for WGAN with $(1, 10^{-10})$-BDP guarantee (our method).}
	\label{tab:accuracy}
	\centering
	\begin{tabular}{ c c c c c }
		\toprule
		{\bf Dataset} 		& {\bf Non-private} 		& {\bf Private classifier} 	& {\bf G-PATE}			& {\bf Our approach} \\
		\midrule
		MNIST 					& $99.20\%$ 				& $95.59\%$					& $56.31\%$ 			& $94.19\%$ \\
		Fashion-MNIST     & $91.51\%$ 					& $80.11\%$						& $51.74\%$ 			& $73.36\%$ \\
		\bottomrule
	\end{tabular}
\end{table*}

We evaluate two major applications of the technique. First, in Section~\ref{sec:debugging}, we show that the generated samples can be used for debugging ML model through data inspection, resembling tasks T1-T4 from~\cite{augenstein2019generative}. Second, we examine the quality of the downstream ML model trained directly on synthetic samples (Section~\ref{sec:learning}), thus demonstrating a possibility of solving T5 (data labelling/annotation) as well.

In the debugging experiment, we attempt to detect a more subtle bug compared to~\cite{augenstein2019generative}: an incorrect image rotation that yields lower model performance. While the pixel intensity inversion can be easily spotted using low-fidelity synthetic samples, image rotation requires higher fidelity to be detected.

Downstream learning experiments are set up as follows:
\begin{enumerate}
\item Train the generative model (\emph{teacher}) on the original data under privacy guarantees.
\item Generate an artificial dataset by the obtained model and use it to train ML models (\emph{students}).
\item Evaluate students on the held-out real test set.
\end{enumerate}

We use two image datasets, MNIST and Fashion-MNIST. Both have $60000$ training and $10000$ test examples, where each example is a $28 \times 28$ size greyscale image. The task of MNIST is handwritten digit recognition, while for Fashion-MNIST it is clothes type recognition. Although these datasets may not be of particular interest from the privacy viewpoint, this choice is defined by the ability to compare to prior work.

Our evaluation is implemented in Python and Pytorch\footnote{\url{http://pytorch.org}}. For the generative model, we experimented with variations of Wasserstein GAN~\cite{arjovsky2017wasserstein} and WGAN-GP~\cite{gulrajani2017improved}, but found the former to produce better results, probably because gradient clipping is already a part of the privacy mechanism. Our critic consists of three convolutional layers with SELU activations~\cite{klambauer2017self} followed by a fully connected linear layer with another SELU and then a linear classifier. The generator starts with a fully connected linear layer that transforms noise (and possibly labels) into a $4096$-dimensional feature vector which is then passed through a SELU activation and three deconvolution layers with SELU activations. The output of the third deconvolution layer is down-sampled by max pooling and normalised with a \texttt{tanh} activation function.

Although we use centralised setting throughout this section, the results are readily transferable to federated scenarios. Previous work suggests that neither GAN sample quality~\cite{triastcyn2019federated} nor BDP guarantees~\cite{triastcyn2019federatedbdp} should be significantly affected.

\begin{figure}
	\centering
    	\includegraphics[width=\linewidth]{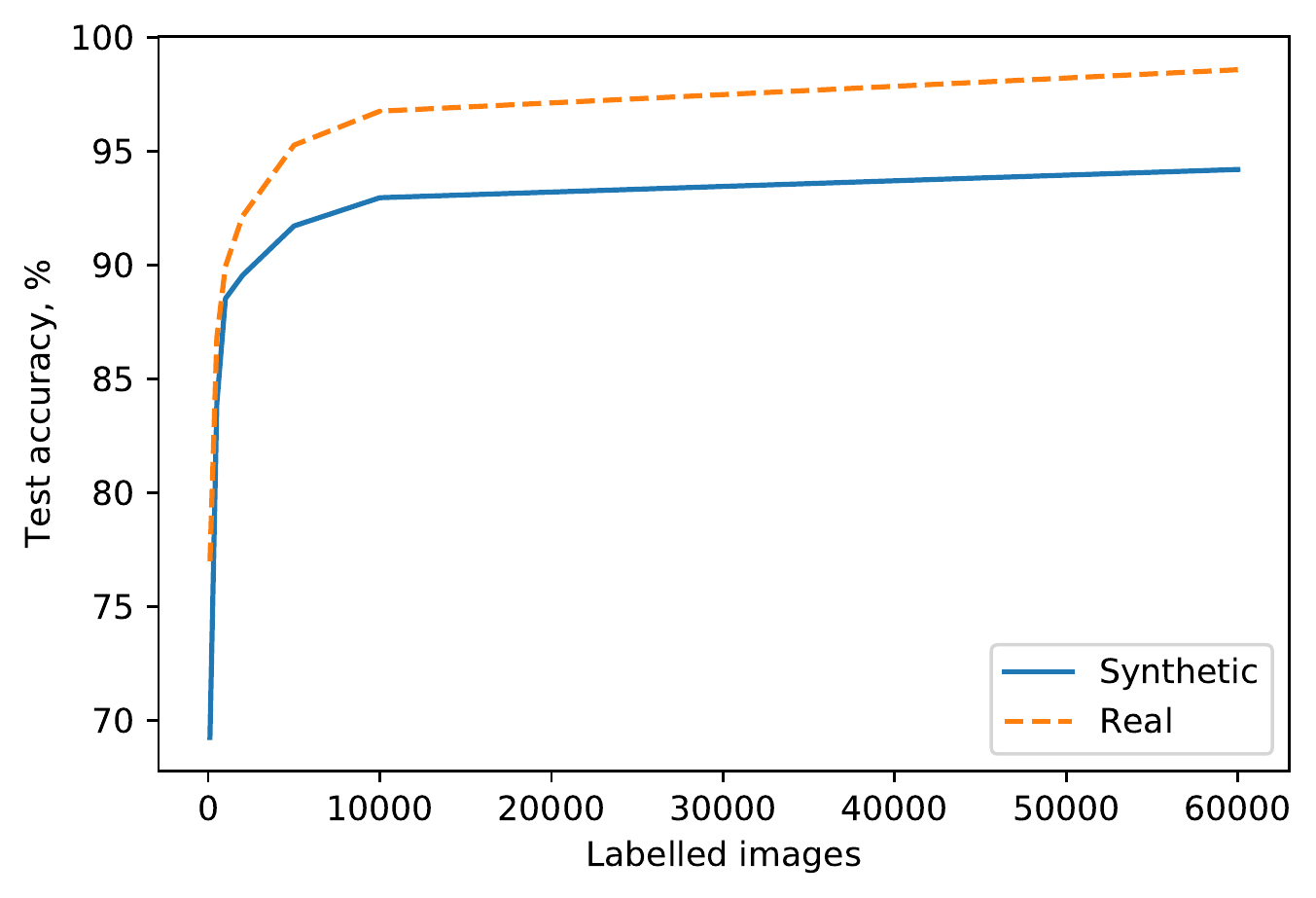}
    	\caption{MNIST accuracy as a function of labelled images.}
    	\label{fig:less_labels_mnist}
\end{figure}

\subsection{Data Inspection}
\label{sec:debugging}

\begin{figure}
	\centering
    	\includegraphics[width=\linewidth]{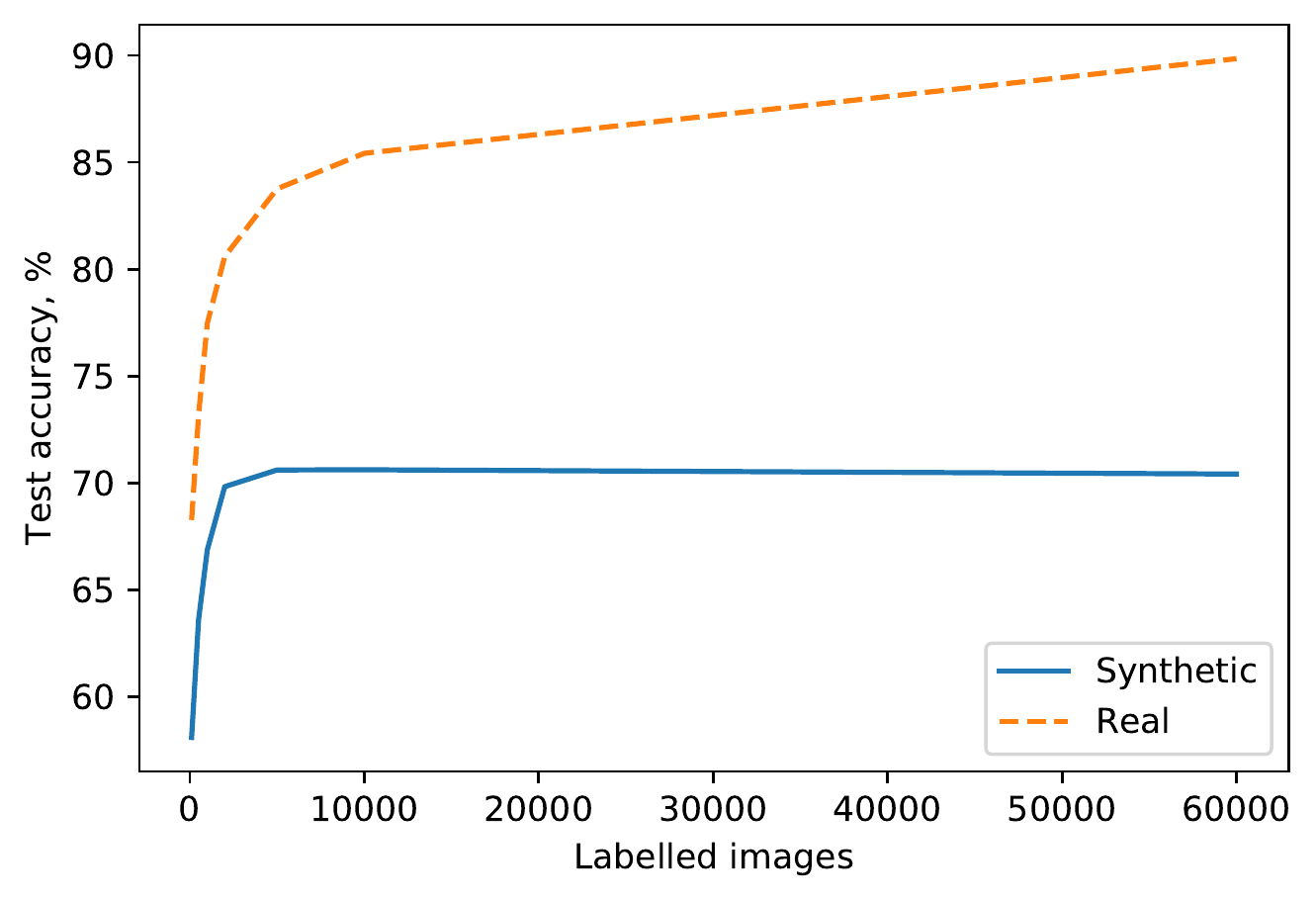}
    	\caption{Fashion-MNIST accuracy as a function of labelled images.}
	\label{fig:less_labels_fashion_mnist}
\end{figure}

The data inspection experiment is setup in the following way. We introduce the rotation bug through randomly rotating some images by $90^\circ$. We then train the two generative models, on correct images and on altered images, and compare their samples. We also train a model with DP to show that its image quality would not be sufficient to detect the error.

Figure~\ref{fig:debug_mnist} shows the output of generative models trained on MNIST with and without image rotation. By examining the samples, developers can clearly determine that a portion of images was rotated. This way, the error can be promptly identified and fixed. On the other hand, with generative models that uphold the traditional DP guarantee (Figure~\ref{fig:debug_mnist_dp}), it would be difficult to detect such pre-processing error, because the produced samples have very low fidelity, even though $\varepsilon$ in this case is unjustifiably high at the order of $10^7$.

We also observe that the synthetic data quality under BDP (see Figures~\ref{fig:fashion_real_fake} and~\ref{fig:debug_mnist_bdp_correct}) is sufficient to detect previously unseen classes or dataset biases, such as under-represented classes. Moreover, these results are achieved with a strong privacy guarantee: under $(1, 10^{-10})$-BDP, and hence, the probability that $(1, 10^{-5})$-DP does not hold for this data is less than $10^{-5}$.

\subsection{Learning Performance}
\label{sec:learning}
Here, we evaluate the generalisation ability of the student model trained on artificial data. More specifically, we train a student model on generated data and report test classification accuracy on a real held-out set.

The goal of this experiment is to show that having a privacy-preserving generative model we can use synthetic samples to fully replace the real data. Not only it allows to eliminate manual labelling of real (and potentially sensitive) data, but also expand the set of problems that can be solved by FL (task T5 in~\citeauthor{augenstein2019generative} classification). For example, some medical data cannot be automatically annotated, and users are not qualified to do that, so high-quality synthetic data would allow the annotation to be performed by doctors without privacy risks for users.

We imitate human annotation by training a separate classifier (with the same privacy guarantee as the generative model) and using it to label synthetic images. While this approach is somewhat different from prior work on generating data for training ML models, comparisons in this section are still valid because our annotator maintains the same privacy guarantee.

We choose to compare with the method called G-PATE~\cite{long2019scalable}, because it is one of the best recent techniques in terms of privacy-utility trade-off. The authors showed that it outperforms another PATE-based approach, PATE-GAN~\cite{jordon2018pate}, as well as DP-GAN~\cite{xie2018differentially}, based on DP-SGD.

Student model accuracy is shown in Table~\ref{tab:accuracy}. Apart from G-PATE, we compare our method to a non-private classifier trained directly on the real dataset, and a private classifier, trained on the real dataset with Bayesian DP. In the case of generative models, the same (non-private) classifier is trained on the private synthetic output. All results in the table are obtained with the privacy guarantee of $(1, 10^{-5})$-DP, or $(1, 10^{-10})$-BDP, which is equivalent to $(1, 10^{-5})$-DP for this data with high probability. Although \cite{long2019scalable} report better results for $(10, 10^{-5})$-DP, we do not include those in the study, because $\varepsilon = 10$ is too high for providing meaningful guarantee~\cite{triastcyn2019bayesian}.

Generally, we observe that on these datasets switching from real to synthetic data does not significantly deteriorate accuracy of the student model while maintaining strong theoretical privacy guarantees. On MNIST, the drop in performance between a private discriminative and a private generative approach is less than $1.5\%$. It is more noticeable on Fashion-MNIST, but is still below $10\%$ and is still lower than the drop between a non-private and a private classifiers. Moreover, as Figures~\ref{fig:less_labels_mnist} and~\ref{fig:less_labels_fashion_mnist} show, models trained on synthetic data achieve sufficiently good performance even when only a small portion of it is labelled. As little as 100 labelled samples is enough to outperform models trained on data generated with comparable DP guarantees.

Interestingly, non-private synthetic data (not shown in the table) allow to reach only marginally better results, suggesting that most of the accuracy loss comes from the generative model rather than privacy preservation. Figures~\ref{fig:less_labels_mnist} and~\ref{fig:less_labels_fashion_mnist} seem corroborate this finding, as synthetic data learning curve quickly saturates.

\section{Conclusions}
\label{sec:conclusion}
We explore the use of generative adversarial networks to tackle the problem of privacy-preserving data inspection and annotation in machine learning. While the previous approaches to this problem involve generative models either without any privacy guarantee or with differential privacy, we opt for a different privacy notion -- Bayesian differential privacy. By capturing the inherent properties of data and allowing for non-uniform privacy loss throughout the dataset, it enables higher-fidelity synthetic data while still maintaining a privacy guarantee comparable to DP.

Our evaluation shows that privacy-preserving GANs with BDP can be used to detect subtle bugs in data itself or pre-processing pipelines, which could not be caught by DP GANs due to low samples fidelity. Similarly, biases in the data and previously unseen classes can be discovered.

In addition, the generated data can be directly annotated and used for training in place of the real data. We demonstrate that student models trained on our synthetic samples achieve significantly higher accuracy compared to prior state-of-the-art and exhibit only a mild drop in performance compared to private classification with real data. Furthermore, this gap is mainly determined by the quality of the generative model, and hence, will get smaller with advances in that field.

\bibliographystyle{named}
\bibliography{ijcai20}

\end{document}